\def\year{2021}\relax
\documentclass[letterpaper]{article} 
\usepackage{aaai21}  
\usepackage{times}  
\usepackage{helvet} 
\usepackage{courier}  
\usepackage[hyphens]{url}  
\usepackage{graphicx} 
\usepackage{natbib}  
\usepackage{caption} 
\usepackage{xcolor}         

\usepackage{mathrsfs} 
\usepackage{amsmath, amssymb, amsfonts}
\usepackage{graphicx}
\usepackage{subcaption}
\usepackage{enumerate}
\usepackage[short,nocomma]{optidef}
\usepackage{booktabs}
\usepackage{multirow}

\usepackage{natbib}
\usepackage{array}
\title{Learning Model Predictive Controllers for Real-Time Ride-Hailing Vehicle Relocation and Pricing Decisions}
\author {
    Enpeng Yuan,
    Pascal Van Hentenryck \\
}
\affiliations {
    Georgia Institute of Technology \\
    eyuan8@gatech.edu, pvh@gatech.edu
}
\begin{document}

\maketitle

\begin{abstract}
Large-scale ride-hailing systems often combine real-time routing at
the individual request level with a macroscopic Model Predictive
Control (MPC) optimization for dynamic pricing and vehicle relocation. The MPC
relies on a demand forecast and optimizes over a longer time horizon
to compensate for the myopic nature of the routing
optimization. However, the longer horizon increases computational complexity and forces the MPC to operate at coarser spatial-temporal granularity,
degrading the quality of its decisions. This paper addresses these
computational challenges by learning the MPC optimization. The
resulting machine-learning model then serves as the optimization proxy
and predicts its optimal solutions. This makes it possible to use
the MPC at higher spatial-temporal
fidelity, since the optimizations can be solved and learned
offline. Experimental results show that the proposed approach improves
quality of service on challenging instances from the New York City dataset.
\end{abstract}

\section{Introduction}

The rapid growth of ride-hailing markets has transformed urban
mobility, offering on-demand mobility services via mobile
applications. While major ride-hailing platforms such as Uber and Didi
leverage centralized dispatching algorithms to find good matching
between drivers and riders, operational challenges persist due
to imbalance between demand and supply. Consider morning rush hours as
an example: most trips originate from residential areas to business
districts where a large number of vehicles accumulate and remain
idle. Relocating these vehicles back to the demand area is thus
crucial to maintaining quality of service and income for the
drivers. In cases where demand significantly exceeds supply, dynamic
pricing is needed to ensure that demand does not surpass service
capacity.

Extensive studies have focused on real-time vehicle relocation and
pricing problems. Existing methodologies fit broadly into two
categories: optimization-based approach and learning-based
approach. Optimization-based approaches involve the solving of a
mathematical program using expected demand and supply information over
a future horizon. Learning-based approaches (predominantly reinforcement
learning) train a state-based decision policy by interacting with
the environment and observing the rewards. While both approaches have
demonstrated promising performance in simulation and (in some cases)
real-world deployment \citep{Didi}, they have obvious drawbacks: the
optimization needs to be solved in real-time and often trades off
fidelity (hence quality of solutions) for computational
efficiency. Reinforcement-learning approaches require a tremendous
amount of data to explore high-dimensional state-action spaces and
often simplify the problem to ensure efficient training. While a
complex real-world system like ride-hailing may never admit a perfect
solution, there are certainly possibilities for improvement.

This paper presents a step to overcoming these computational
challenges. It considers
large-scale ride-hailing systems with real-time routing for
individual requests and a macroscopic Model Predictive Control (MPC) optimization
for dynamic pricing and vehicle relocation.  {\em Its key idea is to
replace the MPC optimization with a machine-learning model that serves
as the optimization proxy and predicts the MPC's optimal solutions.}  The
proposed approach allows ride-hailing systems to consider the MPC at higher spatial or temporal fidelity since the
optimizations can be solved and learned offline.

Learning the MPC however, comes with several challenges. First, the
decisions are interdependent: where the vehicles should relocate
depends on the demand which is governed by price. This imposes
implicit correlation among the predictions, which are hard to enforce
in classic regression models. Second, the predictions are of high
dimensions (e.g., the number of vehicles to relocate between pairs of
zones) and sparse, as relocations typically occur only between a few
low-demand and high-demand regions. Capturing such patterns is
difficult even with large amount of data. Third, the predicted
solutions may not be feasible, as most prediction models cannot
enforce the physical constraints that the solutions need to satisfy.

To solve these challenges, this paper proposes a sequential learning
framework that first predicts the pricing decisions and then the
relocation decisions based on the predicted prices. Furthermore, the
framework utilizes an aggregation-disaggregation procedure
that learns the decisions at an aggregated level to overcome the high
dimensionality and sparsity, and then converts them back to
feasible solutions on the original granularity by a polynomial-time solvable transportation
optimization. \emph{As a
consequence, during real-time operations, the original NP-Hard and
computationally demanding MPC optimization is replaced by a
polynomial-time problem of sequential prediction and optimization}.

The proposed learning \& optimization framework is evaluated on the
New York Taxi data set and serves $6.7\%$ more riders than the
original optimization approach due to its higher fidelity. The results
suggest that {\em a hybrid approach combining machine learning and tractable optimization 
may provide an appealing avenue for certain classes of real-time problems.}

The paper is organized as follows. Section \ref{Sec:prior_works}
summarizes the existing literature. Section \ref{Sec:ride-hail} gives
an overview of the considered real-time ride-hailing operations.
Section \ref{Sec:methodology} contains the main contributions and
presents the learning framework. Section \ref{Sec:expriments}
reports the experimental results on a large-scale case study in New York City.

\section{Related Work}
\label{Sec:prior_works}
While there have been abundant works on the theoretical side of
relocation and pricing (most works model the ride-hailing market as a
one-sided/two-sided market and study properties of different
relocation/pricing policies at market equilibrium), this paper is
interested in real-time pricing and relocation which is reviewed in
this section.

Prior works on real-time pricing and/or relocation fit broadly into
two frameworks: {\em model predictive control (MPC)} (\citep{Miao,
Zhang, Iglesias2017, Huang, ijcai2020} for relocation and \citep{Ma,
Lei-Price} for pricing), and {\em reinforcement learning (RL)}
(\citep{Aug, MOVI, SAMoD, MARL, CoRide, Didi, CP} for relocation
and \citep{Qiu, Lei-Price, Chen-Price} for pricing). MPC is an online
control procedure that repeatedly solves an optimization problem over
a moving time window to find the best control action. System dynamics,
i.e., the interplay between demand and supply, are explicitly modeled
as mathematical constraints. Due to computational complexity, almost
all the MPC models in the literature work at discrete spatial-temporal
scale (dispatch area partitioned into zones, time into epochs) and use
a relatively coarse granularity (a small number of zones or epochs).


Reinforcement learning, on the contrary, does not explicitly model
system dynamics and trains a decision policy offline by approximating
the state/state-action value function. It can be divided
into two streams: single-agent RL and
multi-agent RL. Single-agent RL focuses on
maximizing reward of an individual agent, and multi-agent RL maximizes
collective rewards of all the agents. The main challenge of this approach
is to efficiently learn the state-action value function, which is
high-dimensional (often infinite-dimensional) due to the complex and
fast-changing demand-supply dynamics that arises in real-time
settings. Since RL learns solely from interacting with the environment, a tremendous amount of samples need
to be generated to fully explore the state-action space. Consequently,
many works simplify the problem by using the same policy
for agents within the same region \citep{Aug, MARL}, or restrict
relocations to only neighboring regions \citep{MOVI, SAMoD, CoRide, Didi}. 
In addition, there is no guarantee on the performance of the policy when system dynamics deviates from the environment in which the policy is trained. 

Our approach tries to combine the strength of both worlds - it models
the system dynamics explicitly through a sophisticated MPC model, and
approximates the optimal solutions of the MPC by machine learning to
overcome the real-time computational challenges. As far as the authors
know, the only work that has taken a similar approach is by \citet{Lei},
which learns the decisions of a relocation model and show that the
learned policy performs close to the original model. However, their
model does not include pricing and considers only one epoch (10
mins). This paper focuses on a much more sophisticated MPC
incorporating both relocation and pricing decisions and tracks how
demand and supply interact over the course of multiple epochs. As a
consequence, the model is significantly harder to learn since the
solution space is exponentially larger. To tackle this challenge, this
paper designs an aggregation-disaggregation learning framework and
shows that the learned policy achieves \textbf{superior} performance
than the original model due to its ability to use a finer granularity
within the computational limits.



\section{The Real-Time Ride-Hailing Operations}
\label{Sec:ride-hail}

\begin{figure}[!t]
\centering
\includegraphics[width = \linewidth]{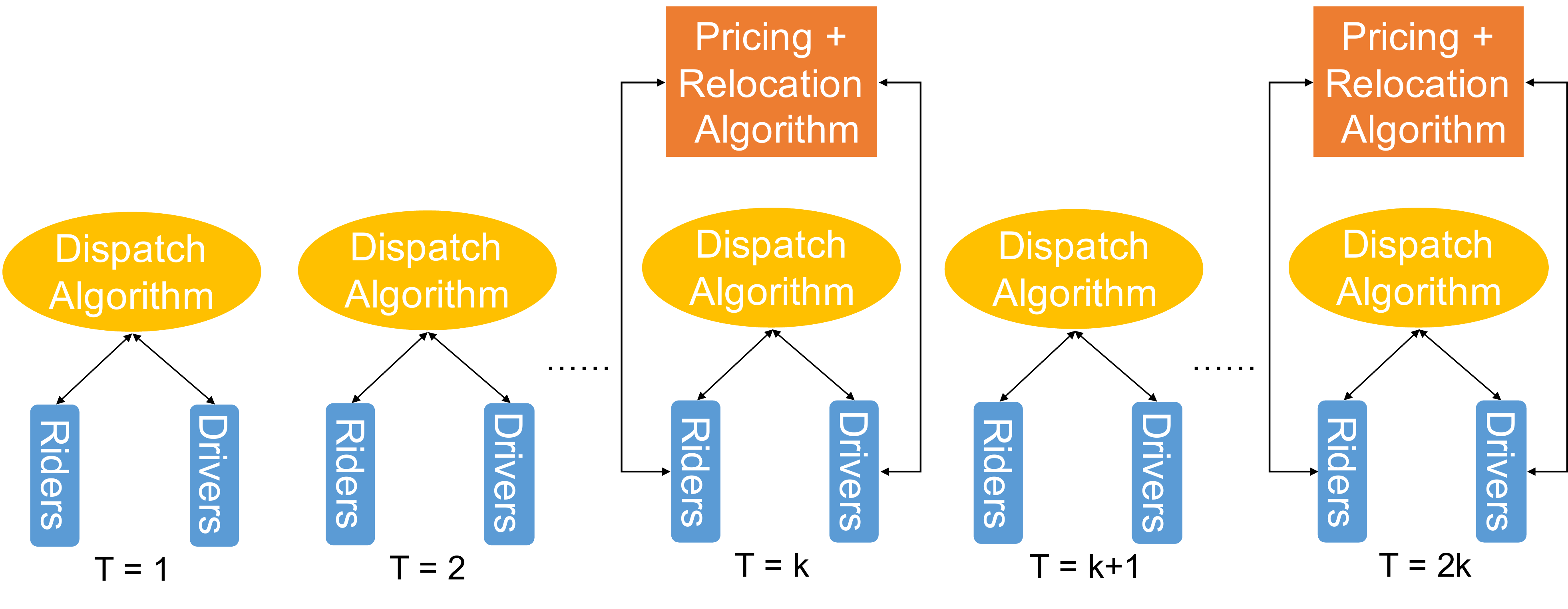}
\caption{The Real-Time Ride-Hailing Operations.}
\label{big-pic}
\end{figure}

This paper considers the real-time ride-hailing framework from \citet{ijcai2020}
(illustrated in Figure \ref{big-pic}). The framework has two key
components: a vehicle routing algorithm and an MPC component for
idle vehicle relocation and pricing. The vehicle routing algorithm
assigns riders to vehicles and chooses the vehicle routes. It operates
at the individual request level with high frequency (e.g., every $30$
seconds). Because of the tight time constraints and the large number
of requests, the routing algorithm is myopic, taking only the
current demand into account. In contrast, the MPC component anticipates the demand and runs at a lower frequency
(e.g., every $5-20$ minutes). The
MPC component tackles the relocation and pricing decisions together as they are interdependent: relocation decisions depend on the demand, and
the demand is shaped by the pricing decisions. The rest of this
section reviews the two components.

\subsection{The Routing Algorithm}
\label{Sec:routing}
The routing algorithm batches requests into a time window and
optimizes every 30 seconds \citep{riley2019}. Its objective is to
minimize a weighted sum of passenger waiting times and penalties for
unserved requests. Each time a request is not scheduled by the routing
optimization, its penalty is increased in the next time window giving
the request a higher priority. The routing algorithm is solved by
column generation: it iterates between solving a restricted master
problem (RMP), which assigns a route (sequence of pickups and
dropoffs) to each vehicle, and a pricing subproblem, which generates
feasible routes for the vehicles. The RMP is depicted in
Figure \ref{fig:master}.  $R$ denotes the set of routes. $V$ the set
of vehicles, and $P$ the set of passengers. $R_v$ denotes the subset
of feasible routes for vehicle $v$. A route is feasible for a vehicle
if it does not exceed the vehicle capacity and does not incur too much
of a detour for its passengers due to ride-sharing. $c_r$ represents
the wait times incurred by all customers served by route $r$. $p_{i}$
is the penalty of not scheduling request $i$, and
$a_{i}^{r}=1$ iff request $i$ is served by route $r$. Decision
variable $y_r \in [0,1]$ is 1 iff route $r$ is selected and $z_{i} \in
[0,1]$ is 1 iff request $i$ is not served by any of the selected
routes. The objective function minimizes the waiting times of the
served customers and the penalties for the unserved
customers. Constraints~\eqref{model:master_constr:allserved} ensure
that $z_{i}$ is set to 1 if request $i$ is not served by any of the
selected routes and constraints~\eqref{model:master_constr:oneroute}
ensure that only one route is selected per vehicle. The column
generation process terminates when the pricing subproblem cannot
generate new routes to improve the solution of the RMP or the solution
time limit is met.

\setcounter{equation}{1}
\begin{figure}[!t]
\begin{subequations} \label{model:master}
\begin{align}
\min 	&  \quad \sum_{r \in R}  c_r y_r + \sum_{i \in P} p_{i} z_{i} \label{model:master_obj}\\
\mbox{s.t.} &  \left ( \sum_{r \in R} y_r a_{i}^{r} \right ) + z_{i} = 1 & \forall i \in P \label{model:master_constr:allserved} \\ 
	&  \sum_{r \in R_v} y_r = 1 & \forall v \in V \label{model:master_constr:oneroute} \\
	&  z_{i} \in \mathbb{N} & \forall i \in P \label{model:master_constr:domainz}\\
	&  y_r \in \{0,1\} & \forall r \in R \label{model:master_constr:domainy}
\end{align}
\end{subequations}
\caption{The Resricted Master Problem Formulation.}
\label{fig:master}
\end{figure}

\subsection{The MPC Component}
\label{Sec:model}


The MPC follows a rolling time horizon approach that discretizes time into epochs of equal length and
performs three tasks at each decision epoch: (1) it predicts the demand for the next $T$
epochs; (2) it optimizes relocation and pricing decisions over these
epochs; and (3) it implements the decisions of the {\em first} epoch
only. Due to the potentially large number of vehicles and riders in
real-time, making pricing and relocation decisions for individual
requests and vehicles is daunting computationally. For this reason,
the MPC component operates at coarser temporal and spatial
granularity: it partitions the geographical area into zones (not
necessarily of equal size or shape) and considers pricing decisions at
the zone level and relocation decisions at the zone-to-zone level. The MPC assumes that vehicles only pick up demand in the same zone and that
vehicles, once they start delivering passengers or relocating, must finish their current trip
before taking another assignment. These assumptions help the MPC
model \emph{approximate} the behavior of the underlying routing algorithm (but
the routing algorithm does not have to obey these constraints). The
only interactions between the routing optimization and the MPC
components are the relocation decisions. To model reasonable waiting
times, riders can only be picked up within $s$ epochs of their
requests: they drop out if waiting more than $s$ epochs.

The impact of pricing on the demand is specified by the following
process. Let $D^0_{ijt}$ denote the number of vehicles needed to
serve expected riders from zone $i$ to zone $j$ under the baseline
price (i.e., without surge pricing or promotion discount). At each
epoch $t$, the MPC component determines the demand multiplier $\gamma_{it}$,
where $\gamma_{it}\in [0,1]$ represents the proportion of demand to
keep in the zone (e.g., $\gamma_{it} = 0.8$ means 80\% demand from zone
$i$ will be kept and 20\% will be priced out). The price corresponding
to each demand multiplier can be estimated from historical market data
and is taken as prior knowledge. The MPC assumes that a set of demand multipliers $\{\gamma^k_{it}\}_{k\in K}$ along with the corresponding demand $\{D^k_{ijt}\}_{k\in K}:= \{\gamma^k_{it}D^0_{ijt}\}_{k\in K}$ is available for each zone and epoch. The model aims at selecting the demand multipliers (and hence the demand) such that every rider is served in reasonable time.



\begin{table}[!t]
    \centering
    \small
    \begin{tabular}{l p{5cm}}
        \midrule
        \multicolumn{2}{c}{Model Input} \\
        \cmidrule{1-2}
        $V_{i t}$ & Number of vehicles that will become idle in $i$ during $t$ \\
        $\{D^k_{ijt}\}_{k\in K}$ & Set of demand from $i$ to $j$ during $t$ available to be selected\\
        $\lambda_{ij}$ & Number of epochs to travel from $i$ to $j$ \\
        \cmidrule{1-2}
        \multicolumn{2}{c}{Model Parameters} \\
        \cmidrule{1-2}
        $s$ & Number of epochs that a rider remains in the system \\
        $W_{ij}$ & Average number of riders from $i$ to $j$ that a vehicle carries  \\
        $q^p(t,\rho)$ & Weight of a rider served at $\rho$ whose request was placed at $t$ \\
        $q^r_{ij}(t)$ & Relocation cost between $i$ and $j$ in $t$ \\
        \cmidrule{1-2} 
        \multicolumn{2}{c}{Decision Variables} \\
        \cmidrule{1-2}
        $p^k_{it} \in \{0,1\}$ & Whether the $k$th demand multiplier $\gamma_{it}^k$ is chosen for zone $i$ and epoch $t$ \\
        $x_{ijt}^r\in \mathbb{Z}_+$ & Number of vehicles starting to relocate from $i$ to $j$ during $t$ \\
        \cmidrule{1-2}
        \multicolumn{2}{c}{Auxiliary Variables} \\
        \cmidrule{1-2}
        $v_{ijt} \in \mathbb{Z}_+$ & Number of vehicles needed to serve all expected riders from $i$ to $j$ whose requests are placed at $t$ \\
        $x_{ijt\rho}^p \in \mathbb{Z}_+$ & Number of vehicles that start to serve at time $\rho$ riders going from $i$ to $j$ whose requests were placed at $t$ \\
        $l_{it} \in \{0,1\}$ & Whether there is unserved demand in $i$ at the end of epoch $t$ \\
        \midrule
    \end{tabular}
    \caption{The Nomenclature for the MPC Optimization.}
    \label{tab:mpc_variables}
\end{table}

\setcounter{equation}{2}
\begin{figure}[t]
\begin{maxi!}[2]<b>
    {}{\underset{i,j}\sum \underset{t,\rho} \sum q^p(t,\rho) W_{ij}x^p_{ijt\rho} - \underset{i,j} \sum \underset{t} \sum q^r_{ij}(t) x^r_{ijt}}
    {}{} \notag
    \addConstraint{\sum_{k\in K} p^k_{it}}{=1}{} \label{constr_single_price} 
    \addConstraint{v_{ijt}}{=\sum_{k\in K} D_{ijt}^k p^k_{it} \quad }{} \label{constr_price_demand}
    \addConstraint{\sum_{\rho\in\phi(t)} x^p_{ijt\rho}}{ = v_{ijt} \;\;\;\;\;\; (t\leq T-s+1)}{} \label{constr_demand1}
    \addConstraint{\sum_{\rho\in\phi(t)} x^p_{ijt\rho}}{ \leq v_{ijt} \;\;\;\;\;\; (t>T-s+1)}{} \label{constr_demand2}
    \addConstraint{\sum_{j,t_0} x^p_{ijt_0 t} + \sum_{j} x^r_{ijt} - V_{i t} =}{}{} \notag
    \addConstraint{\sum_{j,t_0} x^p_{jit_0 (t - \lambda_{ji})} + \sum_{j} x^r_{ji(t - \lambda_{ji})}}{} \label{constr_balance}
    \addConstraint{\sum_{j} x^r_{ijt}}{\leq M l_{it}}{} \label{no_relocate_constraint1}
    \addConstraint{\sum_j \sum_{t_0: \, t\in\phi(t_0)} \left(v_{ijt} - \sum_{\rho=t_0}^{t} x^p_{ijt_0\rho} \right)  }{}{} \notag
    \addConstraint{\leq M(1 - l_{it})}{}{}  \label{no_relocate_constraint2}
    \addConstraint{x^p_{ijt\rho}, x^r_{ijt}, v_{ijt}}{\in \mathbb{Z_+}}{} \label{constr_var_type1}
    \addConstraint{p^k_{it}, l_{it}}{\in \{0,1\}}{} \label{constr_var_type4}
\end{maxi!}
\caption{The MPC Optimization with Pricing \& Relocation.}
\label{fig:MPC}
\end{figure}

\subsubsection{The MPC Formulation}
\label{model-formulation}

The MPC optimization decides the zone pricing and the
zone-to-zone relocation so that every rider is served in $s$ epochs. Table \ref{tab:mpc_variables} summarizes its
nomenclature and Figure \ref{fig:MPC} presents the optimization model. In the model, $i,j$ denote zones and
$t_0,t,\rho$ epochs. The ride-sharing coefficient $W_{ij}$ represents
the average number of riders traveling from $i$ to $j$ that a
vehicle carries accounting for ride-sharing. The time-dependent weights $q^p(t,\rho)$ and
$q^r_{ij}(t)$ are designed to favor serving requests and performing
relocations early, and to avoid postponing them: they are decreasing
in $t$ and $\rho$.

The decision variables $p_{it}$ and $x^r_{ijt}$ capture the pricing and relocation decisions. Although
decisions are made for each epoch in the time
horizon, only the first epoch's decisions are actionable and
implemented: the next MPC execution will reconsider the decisions for
subsequent epochs. Note that the auxiliary variables $x^p_{i j
t \rho}$ are only defined for a subset of the subscripts, since riders
drop out if they are not served in reasonable time. The valid
subscripts for variables $x^p_{i j t\rho}$ must satisfy the constraint
$1 \leq t \leq \rho \leq \min(T, t + s -1)$. These conditions are
implicit in the model for simplicity. Furthermore, $\phi(t)
= \{\rho\in\mathcal{T}: t\leq \rho \leq t+s-1\}$ denotes the set of valid
pick-up epochs for riders placing their requests in epoch $t$.

The MIP model maximizes the weighted sum of customers served and
minimizes the relocation cost. Constraint (\ref{constr_single_price})
ensures that the model selects exactly one demand multiplier for each
zone and epoch. Constraint (\ref{constr_price_demand}) derives the
number of vehicles needed to serve the demand from $i$ to $j$ whose
requests are placed at $t$ as a function of the demand multiplier selected
(captured by variable $p^k_{it}$).  Constraint (\ref{constr_demand1})
enforces the service guarantees: it makes sure that riders with
requests in the first $(T-s+1)$ epochs are served in the time horizon
since they have at least $s$ epochs to be served. Constraint
(\ref{constr_demand2}) makes sure that the served demand does not
exceed the true demand. Constraint (\ref{constr_balance}) is the flow
balance constraint for each zone and epoch. Constraints
(\ref{no_relocate_constraint1}) and (\ref{no_relocate_constraint2})
prevent vehicles from relocating unless all demand in the zone is
served, approximating the behavior of the routing algorithm which
favors scheduling vehicles to nearby requests. Constraints
(\ref{constr_var_type1}) - (\ref{constr_var_type4}) specify the ranges
of the variables.  Constraint (\ref{constr_price_demand}) -
(\ref{constr_demand2}) are defined for all $(i,j,t)$, and
(\ref{constr_single_price}) and (\ref{constr_balance}) - (\ref{no_relocate_constraint2}) are defined
for all $(i,t)$. The model is always feasible when the demand can be
reduced to $0$ in all zones and epochs. The MIP model is challenging
to solve at high fidelity when the number of zones $|Z|$, the length
of time horizon $|\mathcal{T}|$, or the number of demand multipliers
$|K|$ is large. {\em This is the key motivation for learning the MPC
optimization. By solving the MPC optimization offline and learning its
optimal solutions, the MPC component would be able to
operate at finer spatial and temporal granularity and make more
informed decisions.}  This paper explores this avenue.

\section{The Learning Methodology}
\label{Sec:methodology}

\begin{figure}[!t]
\centering
\includegraphics[width = \linewidth]{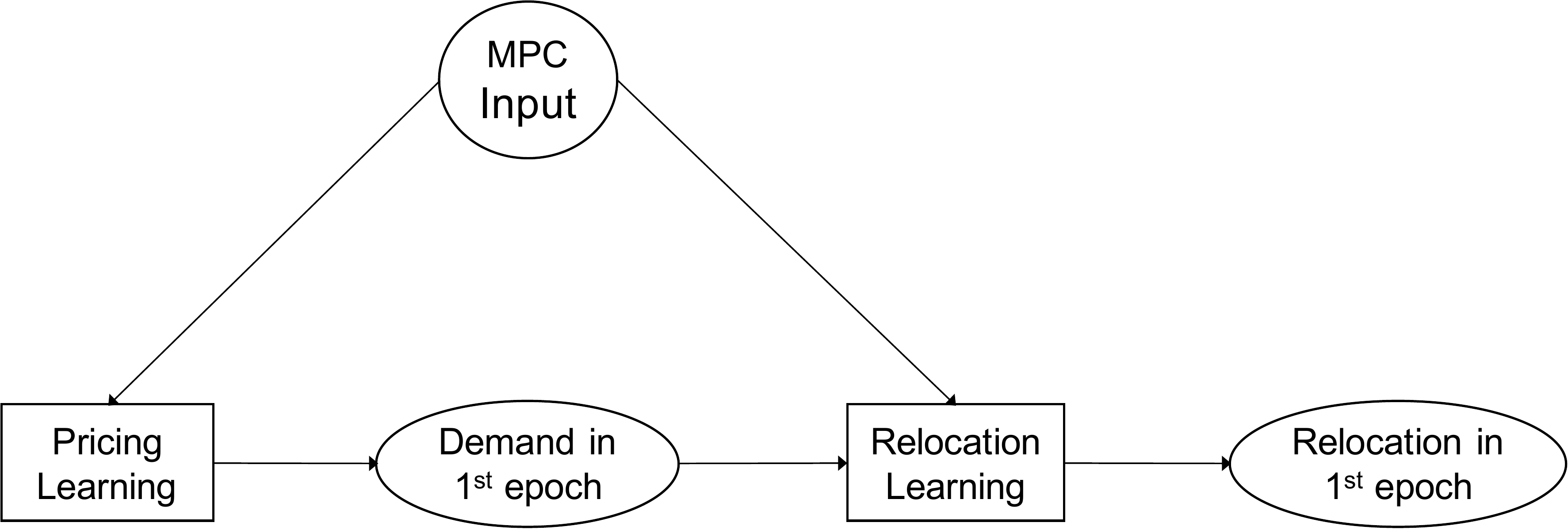}
\caption{The Learning Methodology.}
\label{learning_procedure}
\end{figure}

The learning methodology approximates the MPC decisions in the first epoch since only these decisions are actionable after the MPC execution. Its architecture is depicted in Figure
\ref{learning_procedure}: it first learns the pricing decisions,
which shape the demand, and then the relocation decisions.
This section describes the overall approach. The specifics about the
learning models are given in the experimental results.

\subsection{Learning Pricing Decisions}
\label{Sec:price_learning}

The pricing-learning algorithm takes the same inputs as the MPC
optimization (listed in Table \ref{tab:mpc_variables}) and predicts the demand multipliers $[\gamma_{i1}]_{i\in
Z}$ (and hence the demand) in the first epoch. These predicted multipliers
are rounded to the nearest demand multiplier to be the final pricing
decisions. For example, if the set of demand multipliers are $[1, 0.75, 0.5, 0.25, 0]$ and the prediction for $\gamma_{i1}$ is $0.8$, the final
prediction will be $0.75$.


\subsection{Learning Relocation Decisions}

The relocation decisions are obtained in four steps: aggregation,
learning, feasibility restoration, and disaggregation.

\subsubsection{Aggregation} The relocation decision vector $[x^r_{ij1}]_{i,j\in Z}$
is often sparse, with most relocations occurring between a few
low-demand and high-demand zones. To reduce sparsity, $x^r_{ij1}$ is first aggregated to the zone
level. More precisely, two metrics are learned for each zone $i$: (1) the
number of vehicles relocating from $i$ to other zones, i.e., $y^o_i:=
\sum_{j\in Z, j\neq i} x^r_{ij1}$, and (2) the number of vehicles
relocating to $i$ from other zones, i.e., $y^d_i := \sum_{j\in Z,
 j\neq i} x^r_{ji1}$. These two metrics can be both non-zero at the
 same time: an idle vehicle might be relocated from $i$ to another
 zone for serving a request in the near future, and another vehicle
 could come to $i$ to serve a later request. This aggregation reduces
 the output dimension from $|Z|^2$ to $2|Z|$.

\subsubsection{Learning}

The relocation-learning algorithm takes the MPC inputs and the predicted first-epoch demand from the pricing learning algorithm. It predicts
the aggregated relocation decisions $\mathbf{y} = [y^d_i, y^o_i]_{i\in
Z}$. In general, these predicted decisions are not feasible and violate
the flow balance constraints. 

\subsubsection{Feasibility Restoration}
\label{Sec:rounding}

The feasibility restoration turns the predictions $\mathbf{\hat{y}} =
[\hat{y}^o_i, \hat{y}^d_i]_{i\in Z}$ into feasible relocation
decisions that are integral and obey the (hard) flow balance
constraints. This is performed in three steps. First, $\hat{y}^o_i$
and $\hat{y}^d_i$ are rounded to their nearest non-negative
integers. Second, to make sure that there are not more relocations
than idle vehicles, the restoration sets $\hat{y}^o_i =
\min\{\hat{y}^o_i, V_{i1}\}$. Third, 
$\hat{y}^o_i$ and $\hat{y}^d_i$ must satisfy the flow balance
constraint, e.g., $\sum_{i\in Z} \hat{y}^o_i = \sum_{i\in Z}
\hat{y}^d_i$: this is achieved by setting the two terms to be the
minimum of the two, by randomly decreasing some non-zero elements of
the larger term.

\subsubsection{Disaggregation Through Optimization}
\label{Sec:transport}

The previous steps produce a feasible relocation plan
$[\hat{y}^d_i,\hat{y}^o_i]_{i\in Z}$ at the zone level. The
disaggregation step reconstructs the zone-to-zone relocation via a
transportation optimization. The model formulation is given in Figure
\ref{fig:ASSIGN}. Variable $z_{ij}$ denotes the number of vehicles to
relocate from $i$ to $j$, and constant $c_{ij}$ represents the
corresponding relocation cost. The model minimizes the total
relocation cost to consolidate the relocation plan. The solution
$z_{ij}$ is implemented by the ride-hailing platform in the same way as
$x^r_{ij1}$ from the MPC. Note that $z_{ii}$ should be 0 for all $i$
since $\hat{y}^d_i$ and $\hat{y}^o_i$ denote relocations into and out
of each zone. However, the problem in that form may be infeasible.  By
allowing the $z_{ii}$'s to be positive and assigning a large value to
the relocation costs $c_{ii}$, the problem is always feasible, total
unimodular, and polynomial-time solvable \citep{transportation-TU}.
{\em As a result, the NP-hard MPC optimization is replaced by a
learning methodology that produces an approximate solution in
polynomial time.}

\begin{figure}
\begin{align*}
    \min \quad & \sum_{i,j\in Z} c_{ij} z_{ij} \notag \\
    \mbox{s.t.} \quad & \sum_{j\in Z} z_{ij} = \hat{y}^o_i, \quad & \forall i\in Z\\ 
    & \sum_{j\in Z} z_{ji} = \hat{y}^d_i, \quad & \forall i\in Z\\
    & z_{ij} \in \mathbb{Z^+}, \quad & \forall i,j\in Z
\end{align*}
\caption{The Transportation Problem for Disaggregation.}
\label{fig:ASSIGN}
\end{figure}

\section{Experimental Results}
\label{Sec:expriments}

The learning framework is evaluated on Yellow Taxi Data in Manhattan,
New York City \citep{nycdata}. The learning models are trained from
$2017/01$ to $2017/05$ and tested in $2017/06$. Section \ref{Sec:simu}
reviews the simulation environment, Section \ref{Sec:train} presents
the learning results, and Section \ref{Sec:results} evaluates the performance of the learned policy.

\subsection{Simulation Environment}
\label{Sec:simu}

The end-to-end ride-hailing simulator reviewed in Section \ref{Sec:ride-hail} is the basis of the experimental evaluation. 
The Manhattan area is partitioned into a grid of cells of $200$
squared meter and each cell represents a pickup/dropoff
location. Travel times between the cells are queried
from \citet{OpenStreetMap}. The fleet is fixed to be $1600$ vehicles
with capacity $4$, distributed randomly among the cells at the
beginning of the simulation. Riders must be picked up in 10 minutes
and matched to a vehicle in 5 minutes since their requests, after which
they drop out. The routing algorithm batches requests into a time window and optimizes every 30 seconds.
The MPC component is executed every $5$ minutes. It partitions the Manhattan area into $24$ zones and time into
$5$-minute epochs. The time window contains $6$ epochs and riders
can be served in $2$ epochs following their requests. The number of
idle vehicles in each epoch is estimated by the simulator based on the
current route of each vehicle and the travel times. The ride-share
ratio is $W_{ij}=1.5$ for all $i,j\in Z$. Service weight and
relocation penalty are $q^p(t,\rho)=0.5^{t}0.75^{\rho-t}$ and
$q^r_{ij}(t)=0.001*0.5^{t}\eta_{ij}$ where $\eta_{ij}$ is travel time
between zone $i$ and zone $j$ in seconds. Five demand multipliers
$[1, 0.75, 0.5, 0.25, 0]$ are available for each zone and epoch.

The baseline demand $\mathbf{D}
:= [D^0_{ijt}]_{i,j\in Z, t\in\mathcal{T}}$ for each
origin-destination zone-pair and epoch is forecasted and used to derive demand corresponding to each demand multiplier $\{D^k_{ijt}\}_{k\in K}$. The design of demand forecasting is beyond the scope of this paper, but the next paragraphs briefly review the 
methodology. In the real world, the
zone-to-zone demand within a short time horizon (e.g., 5-minute time
interval) is highly sparse: most trips travel between a few hot spots. $\mathbf{D}$ is also
high dimensional and has $|Z|^2$ entries. To tackle the challenges
raised by high dimensionality and sparsity, the demand is first
aggregated and predicted at the zone level. Once the zone-level demand
is predicted, it is disaggregated to the zone-to-zone level based on
historical trip destination distribution. For example, if $\mu_{ij}$ proportion of trips from zone $i$ goes to zone $j$, and
$D_{it}$ is the demand prediction for zone $i$, the final zone-to-zone
prediction $D^0_{ijt}$ is $D_{it} \times \mu_{ij}$ rounded to the
nearest integer.

\begin{figure}[!t]
    \begin{subfigure}{0.5\linewidth}
    \includegraphics[width=\linewidth]{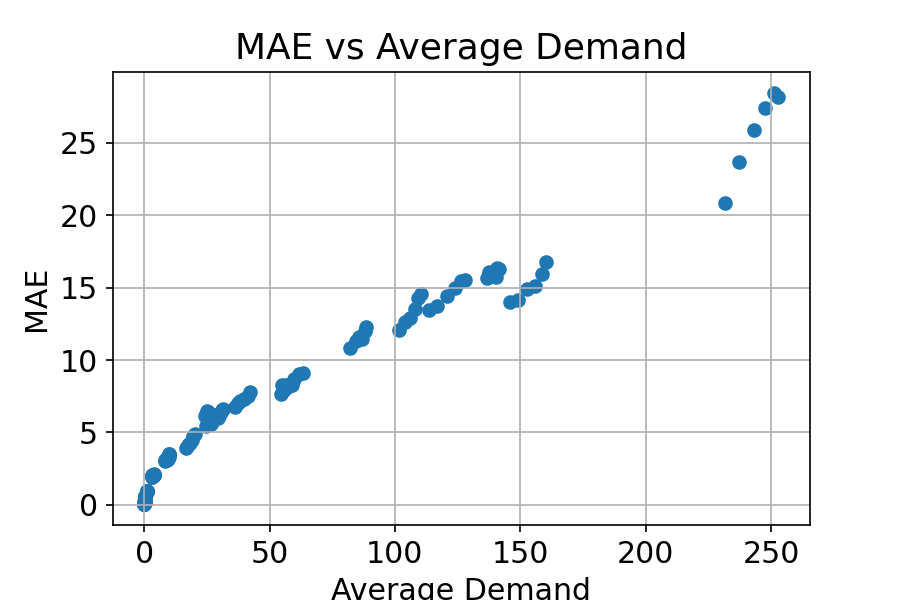}
    \subcaption{MAE.}
    \label{fig:pred_mae}
    \end{subfigure}%
    \begin{subfigure}{0.5\linewidth}
    \includegraphics[width=\linewidth]{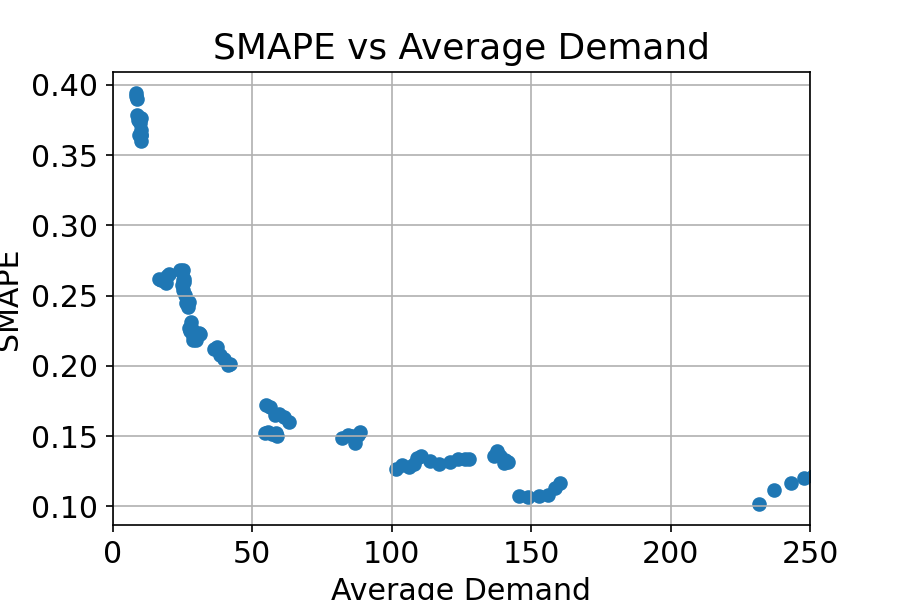}
    \subcaption{SMAPE.}
    \label{fig:pred_smape}
    \end{subfigure}%
    \caption{The Accuracy of the Demand Forecasting. Each Point Denotes the Average Demand and the MAE/SMAPE of a Zone on the Test Set.}
    \label{fig:pred_error}
\end{figure}

The zone-level demand is forecasted based on the zone demand observed
in the last $k$ epochs and the zone demand observed in the last week
for the same periods. For instance, when forecasting the demand
between 7.55 am and 8.00 am on $2017/01/08$, the model uses the demand
between 7.55 am and 8.00 am on $2017/01/01$. The forecasting model is an artificial neural
network with mean squared error (MSE) loss and $l_1$ regularization,
two fully-connected layers with $(256, 256)$ hidden units and ReLU
activation functions. It is trained on $2017/01 - 2017/05$ and tested on
$2017/06$. The mean absolute error (MAE) and the symmetric mean absolute
percentage error (SMAPE) for each zone on the test set are displayed
in Figure \ref{fig:pred_error}. The mean squared error of the
disaggregated zone-to-zone level prediction is 0.49.

The MPC's pricing decisions are
implemented at the level of demand multipliers: if MPC
decides to keep $50\%$ demand in a zone, the simulation randomly
selects $50\%$ requests in the current epoch and discards the rest. After
the MPC decides zone-to-zone level relocations, a vehicle assignment
optimization determines which individual vehicles to relocate by
minimizing total traveling distances (see \citet{ijcai2020} for details).

Of the routing, MPC, and vehicle assignment models, the routing model is the most computationally
intensive since it operates at the individual (driver and rider) level
as opposed to the zone level. Since all three models must be executed
within 30 seconds, the platform allocates $20$ seconds to the routing
optimization, $5$ seconds to the MPC, and $5$ seconds to the vehicle
assignment. All the models are solved using Gurobi 9.0 with 6 cores of
2.1 GHz Intel Skylake Xeon CPU \citep{gurobi}.


\subsubsection{Training Data}

The learning policy is trained on Yellow Taxi data between $2017/01$ and $2017/05$. Each daily instance between 7
am and 9 am are selected as training instances. The total number of
riders in these instances ranges from 10,000 to 50,000, representing a
wide variety of demand scenarios in Manhattan. The instances are then
perturbed by randomly adding/deleting certain percentage of requests
to generate more training instances, where the percentages are sampled
from a uniform distribution $U(-5, 5)$. The instances are run by the
simulator and the MPC results are extracted as training data.

\subsection{Learning Results}
\label{Sec:train}
In
addition to the input features mentioned in
Section \ref{Sec:price_learning}, the pricing-learning algorithm also uses the
following features: the difference between supply and demand
in each zone in the first epoch under all demand multipliers, and the
ratio between cumulative supply and cumulative baseline demand
$[\Tilde{V}_{it} / \Tilde{D}^0_{it}]_{i\in Z, t\in\mathcal{T}}$ where $\Tilde{V}_{it} = \sum_{\tau=1}^t V_{i\tau}$ and
$\Tilde{D}^0_{it} = \sum_{\tau=1}^t D^0_{i\tau}$. These additional
features help improve the learning accuracy.

Four models were trained to learn the pricing and relocation
decisions: random forest (RF), support vector regression (SVR),
gradient boosting regression tree (GBRT), and deep neural network
(DNN). The RF, SVR, and GBRT models were trained on 8000 data points and the DNN was trained on 35000 data points since fitting the DNN typically requires more
data. All the models use the mean squared error (MSE) loss except SVR
which uses epsilon-insensitive loss and $l_2$ regularization. The
hyperparameters of each model were tuned through 5-fold
cross-validation. The selected hyperparameters for the relocation
models are: (kernel, regularization weight) = (radial basis
function, 100), (max tree depth, number of trees) = (32, 200) for RF,
(max tree depth, number of trees) = (64, 200) for GBRT, and two
fully-connected hidden layers with (750, 1024) hidden units and
hyperbolic tangent (tanh) activation functions for the DNN. The
selected hyperparameters for the pricing models are: (kernel, regularization weight) = (radial basis function, 1000), (max tree depth,
number of trees) = (64, 200) for RF, (max tree depth, number of trees)
= (32, 100) for GBRT, and two fully-connected hidden layers with (750,
1024) hidden units and ReLU activation functions for DNN. The SVR,
GBRT, and RF models were trained in scikit-learn package and the DNN
model was trained in Pytorch by Adam optimizer with batch size $32$
and learning rate $10^{-3}$ \citep{sklearn, Adam, PyTorch}.


\begin{table}[!t]
    \centering
    \resizebox{\linewidth}{!} {%
    \begin{tabular}{l c c c}
        \toprule
        Model & Relocation (MSE) & Pricing (MSE) & Pricing (0-1 Loss(\%))\\
        \midrule
        SVR & 21.98 & 100.49 & 13.17  \\
        RF & 21.69 & 101.29 & 13.32  \\
        GBRT & 28.37 & 88.59 & 12.28   \\
        DNN & \textbf{6.83} & \textbf{66.65} & \textbf{9.46} \\
        \bottomrule
    \end{tabular}}
    \caption{Element-Wise Validation Errors.}
    \label{tab:val_error}
\end{table}

\begin{figure}[!t]
    \begin{subfigure}{0.5\linewidth}
    \includegraphics[width=\linewidth]{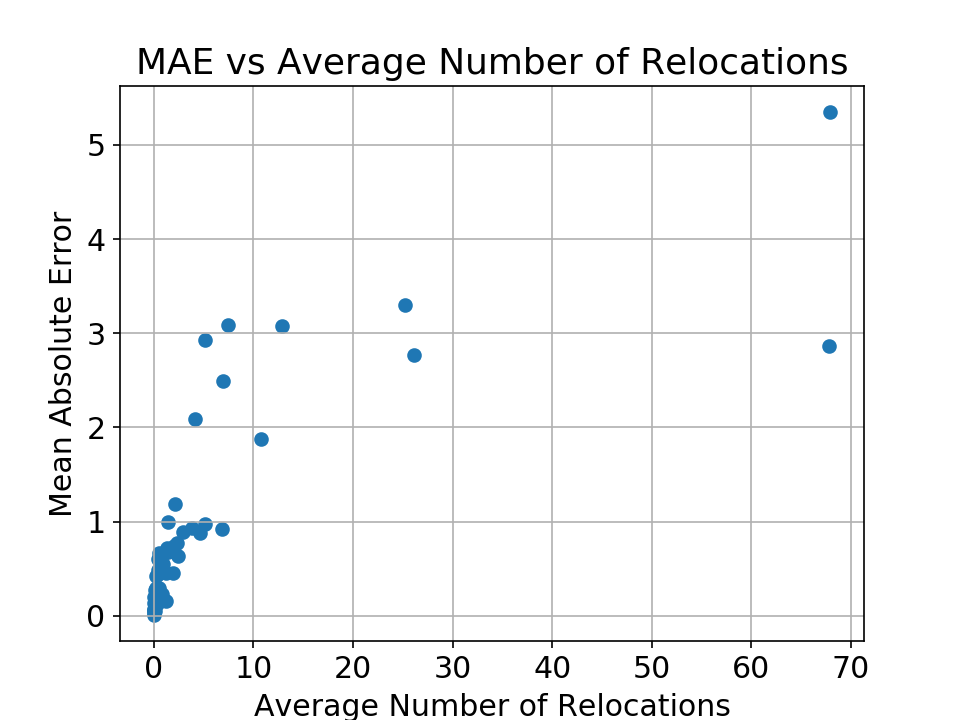}
    \subcaption{Relocation Predictions.}
    \label{fig:reloc_mae}
    \end{subfigure}\hfill
    \begin{subfigure}{0.5\linewidth}
    \includegraphics[width=\linewidth]{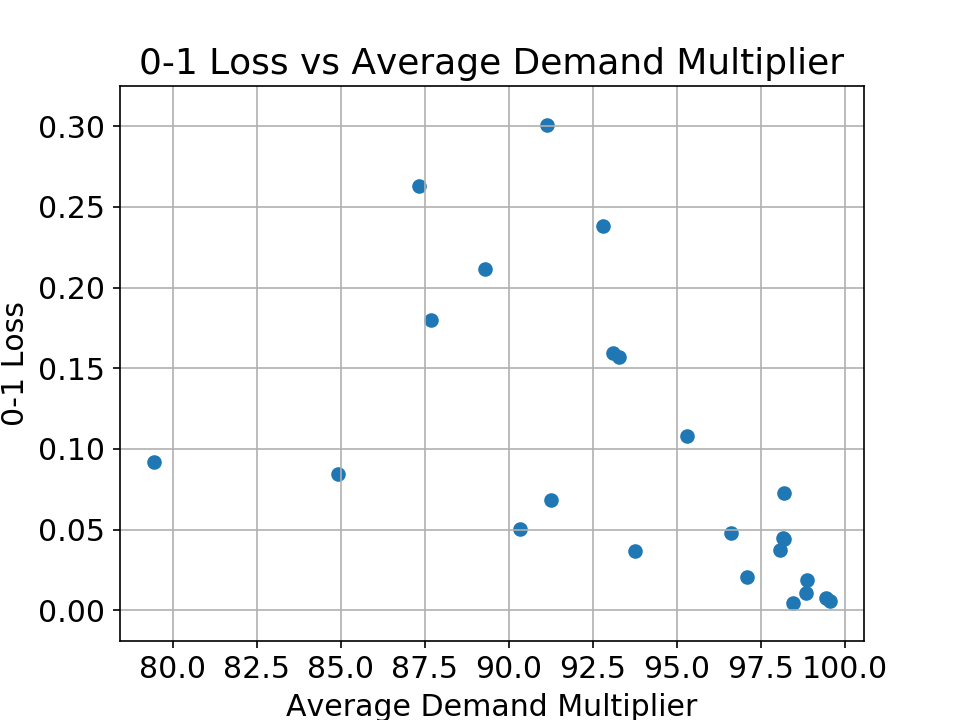}
    \subcaption{Pricing Predictions.}
    \label{fig:price_0-1}
    \end{subfigure}\hfill
    \caption{MPC Decisions Predictions: Each Point on the Left Denotes the Average Relocations and the MAE of Relocation Predictions for a Zone. Each Point on the Right Denotes the Average Demand Multiplier and the 0-1 Loss of Pricing Predictions of a Zone.}
    \label{fig:element_error}
\end{figure}

The trained models were evaluated on a holdout validation set. The predictions are rounded to feasible
solutions by the procedures described in \ref{Sec:price_learning}
and \ref{Sec:rounding}. Element-wise errors after rounding are
reported in Table \ref{tab:val_error}, where the relocation models
report the mean squared error loss (MSE) and the pricing models report
both the MSE and the 0-1 loss (percentage of time that the rounded
predictions were wrong). Since the DNN models achieved the highest
accuracy in both cases, they were selected as the final model. The error for each zone under the DNN model is given in Figure \ref{fig:element_error}. The
prediction errors for all zones are reasonable, although a few
zones exhibit higher loss than others. Overall these results
indicate that the models successfully learned the MPC decisions.

\subsection{The Benefits of Learning the MPC Model}
\label{Sec:results}

\begin{figure}[!t]
    \begin{subfigure}[b]{0.5\linewidth}
    \includegraphics[width=\linewidth]{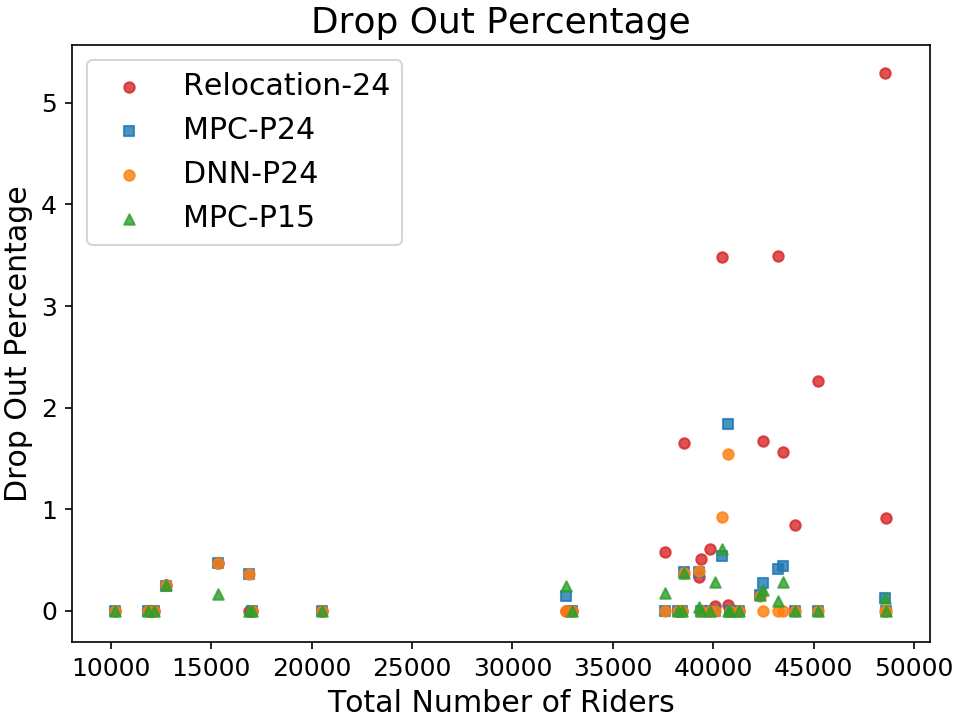}
    \subcaption{Drop Out Percentage.}
    \label{fig1a}
    \end{subfigure}\hfill
    \begin{subfigure}[b]{0.5\linewidth}
    \includegraphics[width=\linewidth]{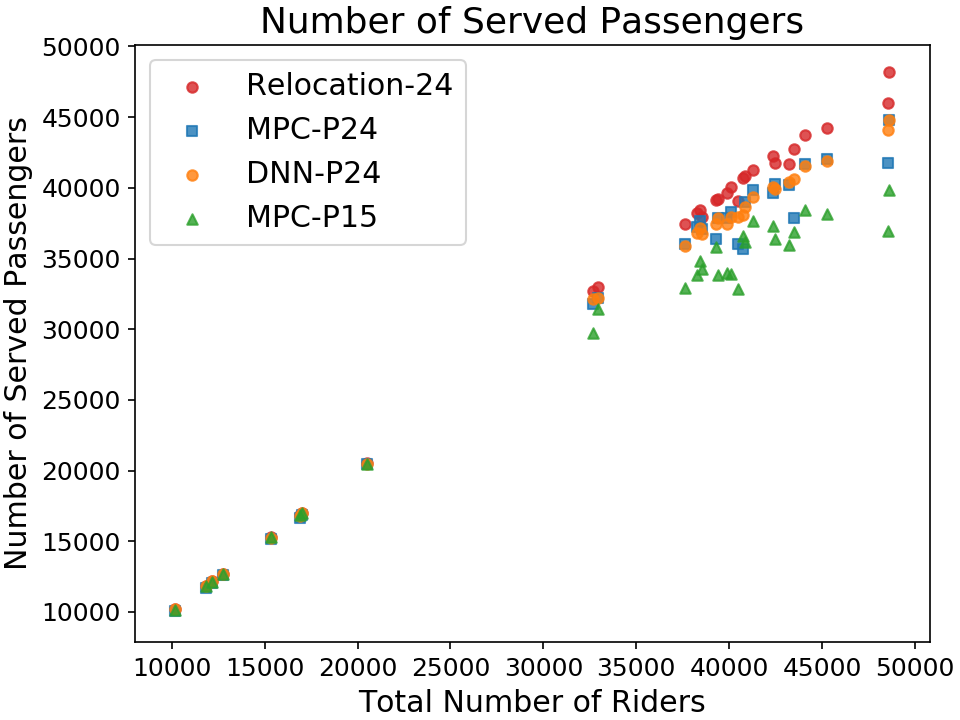}
    \subcaption{Riders Served.}
    \label{fig1b}
    \end{subfigure}\hfill
    \begin{subfigure}[b]{0.5\linewidth}
    \includegraphics[width=\linewidth]{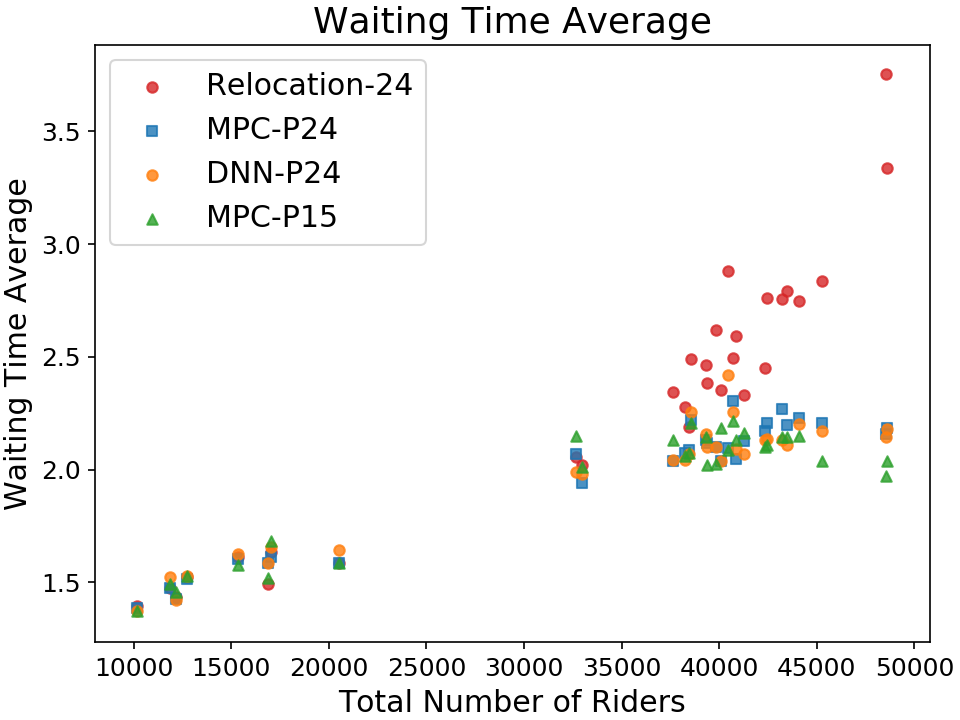}
    \subcaption{Waiting Time Averages.}
    \label{fig1c}
    \end{subfigure}\hfill
    \begin{subfigure}[b]{0.5\linewidth}
    \includegraphics[width=\linewidth]{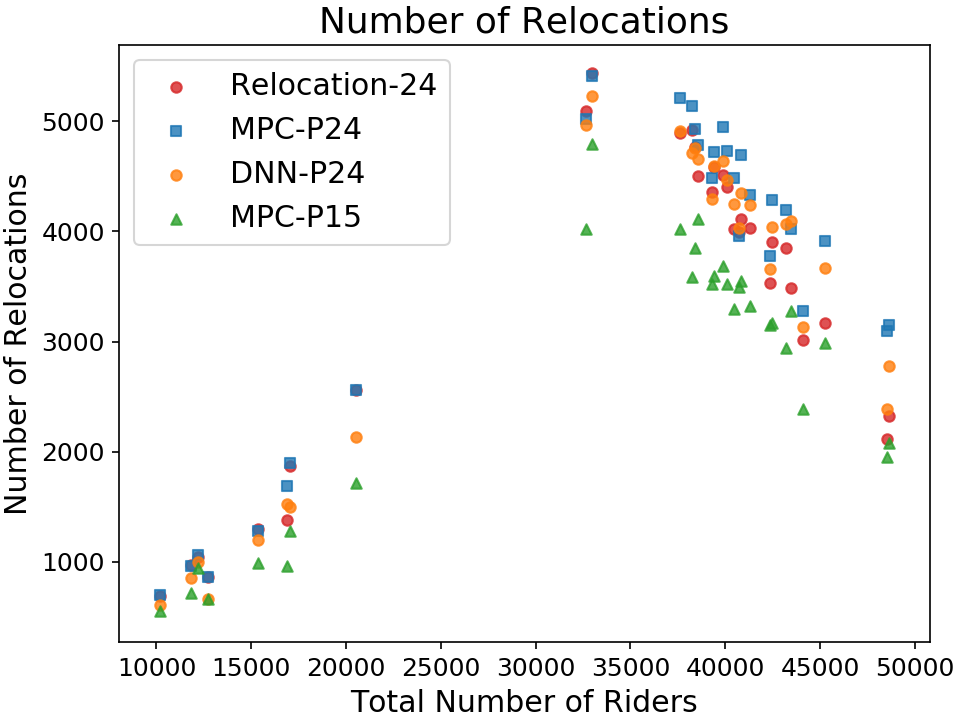}
    \subcaption{Number of Relocations.}
    \label{fig1d}
    \end{subfigure}\hfill
    \caption{Evaluation Results of DNN-P24 and Its Comparison With Pure Optimization Approaches.}
    \label{fig1}
\end{figure}


The benefits of learning the MPC optimization are evaluated on Yellow
Taxi data in June, 2017. The proposed methodology (DNN-P24) is
compared with the original MPC model with $24$ zones (MPC-P24), a
lower-fidelity MPC model with $15$ clustered zones that represents
what can be solved within the computational limit (MPC-P15), and a
baseline that only performs relocation but not pricing
(Relocation-24). {\em All models but MPC-P24 are solved near
optimally in $5$ seconds with 6 CPU cores as discussed
in \ref{Sec:simu}. MPC-P24, which cannot be solved near-optimally in
$5$ seconds, is given more time and represents the ideal solution that
cannot be achieved in real-time.} In particular, 2.4\% MPC-P24  instances failed to find a solution within 20\% optimality gap in 5 seconds. The drop-out rate,
number of riders served, rider waiting time averages, and the number
of relocations are reported in Figure \ref{fig1}. In all instances,
DNN-P24 achieves similar performance as MPC-P24. Both approaches
ensure a drop-out rate near zero, whereas Relocation-24 loses
increasingly more riders as the instance becomes larger. DNN-P24 and
MPC-P24 also achieve lower average waiting times and serve similar
number of riders as Relocation-24. In addition, they serve more riders
than MPC-P15: on large instances with more than 25,000 riders, DNN-P24
serves on average $6.7\%$ more riders than MPC-P15 by pricing out fewer riders, demonstrating the
benefits of higher model fidelity. DNN-P24 and MPC-P24 perform
more relocations than MPC-P15 to serve more riders (finer spatial
partition also unveils more opportunities for vehicle relocation). The
solver times of the transportation optimization never exceed 0.093
seconds and the prediction time is also within fraction of a second. {\em
Overall, these promising results demonstrate that the proposed
framework is capable of approximating high-fidelity MPC model efficiently, which
leads to significant improvements in service quality.}

\section{Conclusion}

Large-scale ride-hailing systems often combine real-time routing at
the individual request level with a macroscopic Model Predictive
Control (MPC) optimization for dynamic pricing and vehicle relocation. The MPC operates over a longer time horizon
to anticipate the demand. The longer horizon increases computational complexity and forces
the MPC to use coarser spatial-temporal granularity, which
degrades the quality of its decisions. This paper addresses
this computational challenge by learning the MPC optimization. The
resulting machine-learning model serves as the optimization proxy, which makes it possible to use
the MPC at higher spatial and/or temporal
fidelity since the optimizations can be solved and learned
offline. Experimental results on the New York Taxi data set show that
the proposed approach serves 6.7\% more riders than the original
optimization approach due to its higher fidelity. A key factor behind
the MPC model as well as the learning is the ability to predict
future demand and supply accurately, which is challenging due to the
fast-changing and volatile nature of real-time dynamics. Therefore,
future works can focus on developing end-to-end learning and
optimization systems that are robust to input uncertainty, possibly
leveraging stochastic optimization and robust training techniques.

\bibliography{aaai}

\begin{thebibliography}{25}
\providecommand{\natexlab}[1]{#1}
\providecommand{\url}[1]{\texttt{#1}}
\providecommand{\urlprefix}{URL }
\expandafter\ifx\csname urlstyle\endcsname\relax
  \providecommand{\doi}[1]{doi:\discretionary{}{}{}#1}\else
  \providecommand{\doi}{doi:\discretionary{}{}{}\begingroup
  \urlstyle{rm}\Url}\fi

\bibitem[{Chen et~al.(2019)Chen, Jiao, Qin, Tang, Li, An, Zhu, and
  Ye}]{Chen-Price}
Chen, H.; Jiao, Y.; Qin, Z.; Tang, X.; Li, H.; An, B.; Zhu, H.; and Ye, J.
  2019.
\newblock InBEDE: Integrating Contextual Bandit with TD Learning for Joint
  Pricing and Dispatch of Ride-Hailing Platforms.
\newblock In \emph{2019 IEEE International Conference on Data Mining (ICDM)},
  61--70.
\newblock \doi{10.1109/ICDM.2019.00016}.

\bibitem[{Gu{\'e}riau and Dusparic(2018)}]{SAMoD}
Gu{\'e}riau, M.; and Dusparic, I. 2018.
\newblock SAMoD: Shared Autonomous Mobility-on-Demand using Decentralized
  Reinforcement Learning.
\newblock \emph{2018 21st International Conference on Intelligent
  Transportation Systems (ITSC)} 1558--1563.

\bibitem[{{Gurobi~Optimization}(2020)}]{gurobi}
{Gurobi~Optimization}. 2020.
\newblock Gurobi Optimizer Reference Manual.
\newblock \urlprefix\url{http://www.gurobi.com}.

\bibitem[{Huang, de~Almeida~Correia, and An(2018)}]{Huang}
Huang, K.; de~Almeida~Correia, G.~H.; and An, K. 2018.
\newblock Solving the station-based one-way carsharing network planning problem
  with relocations and non-linear demand.
\newblock \emph{Transportation Research Part C: Emerging Technologies} 90:
  1--17.
\newblock ISSN 0968-090X.
\newblock \doi{https://doi.org/10.1016/j.trc.2018.02.020}.
\newblock
  \urlprefix\url{https://www.sciencedirect.com/science/article/pii/S0968090X18302511}.

\bibitem[{Iglesias et~al.(2017)Iglesias, Rossi, Wang, Hallac, Leskovec, and
  Pavone}]{Iglesias2017}
Iglesias, R.; Rossi, F.; Wang, K.; Hallac, D.; Leskovec, J.; and Pavone, M.
  2017.
\newblock Data-Driven Model Predictive Control of Autonomous Mobility-on-Demand
  Systems.
\newblock \emph{CoRR} abs/1709.07032.
\newblock \urlprefix\url{http://arxiv.org/abs/1709.07032}.

\bibitem[{Jiao et~al.(2021)Jiao, Tang, Qin, Li, Zhang, Zhu, and ping Ye}]{Didi}
Jiao, Y.; Tang, X.; Qin, Z.; Li, S.; Zhang, F.; Zhu, H.; and ping Ye, J. 2021.
\newblock Real-world Ride-hailing Vehicle Repositioning using Deep
  Reinforcement Learning.
\newblock \emph{ArXiv} abs/2103.04555.

\bibitem[{Jin et~al.(2019)Jin, Zhou, Zhang, Li, Guo, Qin, Jiao, Tang, Wang,
  Wang, and et~al.}]{CoRide}
Jin, J.; Zhou, M.; Zhang, W.; Li, M.; Guo, Z.; Qin, Z.; Jiao, Y.; Tang, X.;
  Wang, C.; Wang, J.; and et~al. 2019.
\newblock CoRide: Joint Order Dispatching and Fleet Management for Multi-Scale
  Ride-Hailing Platforms.
\newblock \emph{Proceedings of the 28th ACM International Conference on
  Information and Knowledge Management} \doi{10.1145/3357384.3357978}.
\newblock \urlprefix\url{http://dx.doi.org/10.1145/3357384.3357978}.

\bibitem[{Kingma and Ba(2014)}]{Adam}
Kingma, D.; and Ba, J. 2014.
\newblock Adam: A Method for Stochastic Optimization.
\newblock \emph{International Conference on Learning Representations} .

\bibitem[{Lei, Jiang, and Ouyang(2019)}]{Lei-Price}
Lei, C.; Jiang, Z.; and Ouyang, Y. 2019.
\newblock Path-based dynamic pricing for vehicle allocation in ridesharing
  systems with fully compliant drivers.
\newblock \emph{Transportation research procedia} 38: 77--97.

\bibitem[{Lei, Qian, and Ukkusuri(2020)}]{Lei}
Lei, Z.; Qian, X.; and Ukkusuri, S.~V. 2020.
\newblock Efficient proactive vehicle relocation for on-demand mobility service
  with recurrent neural networks.
\newblock \emph{Transportation Research Part C: Emerging Technologies} 117:
  102678.
\newblock ISSN 0968-090X.
\newblock \doi{https://doi.org/10.1016/j.trc.2020.102678}.
\newblock
  \urlprefix\url{https://www.sciencedirect.com/science/article/pii/S0968090X20305933}.

\bibitem[{Liang et~al.(2021)Liang, Wen, Lam, Sumalee, and Zhong}]{CP}
Liang, E.; Wen, K.; Lam, W.; Sumalee, A.; and Zhong, R. 2021.
\newblock An Integrated Reinforcement Learning and Centralized Programming
  Approach for Online Taxi Dispatching.
\newblock \emph{IEEE Transactions on Neural Networks and Learning Systems}
  \doi{10.1109/TNNLS.2021.3060187}.

\bibitem[{Lin et~al.(2018)Lin, Zhao, Xu, and Zhou}]{MARL}
Lin, K.; Zhao, R.; Xu, Z.; and Zhou, J. 2018.
\newblock Efficient Large-Scale Fleet Management via Multi-Agent Deep
  Reinforcement Learning.
\newblock In \emph{Proceedings of the 24th ACM SIGKDD International Conference
  on Knowledge Discovery and Data Mining}, KDD '18, 1774–1783. New York, NY,
  USA: Association for Computing Machinery.
\newblock ISBN 9781450355520.
\newblock \doi{10.1145/3219819.3219993}.
\newblock \urlprefix\url{https://doi.org/10.1145/3219819.3219993}.

\bibitem[{Ma, Fang, and Parkes(2019)}]{Ma}
Ma, H.; Fang, F.; and Parkes, D.~C. 2019.
\newblock Spatio-Temporal Pricing for Ridesharing Platforms.
\newblock In \emph{Proceedings of the 2019 ACM Conference on Economics and
  Computation}, EC '19, 583. New York, NY, USA: Association for Computing
  Machinery.
\newblock ISBN 9781450367929.
\newblock \doi{10.1145/3328526.3329556}.
\newblock \urlprefix\url{https://doi.org/10.1145/3328526.3329556}.

\bibitem[{Miao et~al.(2015)Miao, Lin, Munir, Stankovic, Huang, Zhang, He, and
  Pappas}]{Miao}
Miao, F.; Lin, S.; Munir, S.; Stankovic, J.; Huang, H.; Zhang, D.; He, T.; and
  Pappas, G. 2015.
\newblock Taxi Dispatch with Real-Time Sensing Data in Metropolitan Areas — a
  Receding Horizon Control Approach.
\newblock \emph{IEEE Transactions on Automation Science and Engineering} 13.
\newblock \doi{10.1145/2735960.2735961}.

\bibitem[{NYC(2019)}]{nycdata}
NYC. 2019.
\newblock NYC Taxi \& Limousine Commission - Trip Record Data.
\newblock
  \urlprefix\url{http://www.nyc.gov/html/tlc/html/about/trip\_record\_data.shtml}.
\newblock Accessed: 2020-10-01.

\bibitem[{Oda and Joe-Wong(2018)}]{MOVI}
Oda, T.; and Joe-Wong, C. 2018.
\newblock MOVI: A Model-Free Approach to Dynamic Fleet Management.
\newblock \emph{IEEE INFOCOM 2018 - IEEE Conference on Computer Communications}
  2708--2716.

\bibitem[{OpenStreetMap(2017)}]{OpenStreetMap}
OpenStreetMap. 2017.
\newblock Planet dump retrieved from https://planet.osm.org.
\newblock \urlprefix\url{https://www.openstreetmap.org}.
\newblock Accessed: 2020-10-01.

\bibitem[{Paszke et~al.(2019)Paszke, Gross, Massa, Lerer, Bradbury, Chanan,
  Killeen, Lin, Gimelshein, Antiga, Desmaison, Kopf, Yang, DeVito, Raison,
  Tejani, Chilamkurthy, Steiner, Fang, Bai, and Chintala}]{PyTorch}
Paszke, A.; Gross, S.; Massa, F.; Lerer, A.; Bradbury, J.; Chanan, G.; Killeen,
  T.; Lin, Z.; Gimelshein, N.; Antiga, L.; Desmaison, A.; Kopf, A.; Yang, E.;
  DeVito, Z.; Raison, M.; Tejani, A.; Chilamkurthy, S.; Steiner, B.; Fang, L.;
  Bai, J.; and Chintala, S. 2019.
\newblock PyTorch: An Imperative Style, High-Performance Deep Learning Library.
\newblock In Wallach, H.; Larochelle, H.; Beygelzimer, A.; d\textquotesingle
  Alch\'{e}-Buc, F.; Fox, E.; and Garnett, R., eds., \emph{Advances in Neural
  Information Processing Systems 32}, 8024--8035. Curran Associates, Inc.
\newblock
  \urlprefix\url{http://papers.neurips.cc/paper/9015-pytorch-an-imperative-style-high-performance-deep-learning-library.pdf}.

\bibitem[{Pedregosa et~al.(2011)Pedregosa, Varoquaux, Gramfort, Michel,
  Thirion, Grisel, Blondel, Prettenhofer, Weiss, Dubourg, Vanderplas, Passos,
  Cournapeau, Brucher, Perrot, and Duchesnay}]{sklearn}
Pedregosa, F.; Varoquaux, G.; Gramfort, A.; Michel, V.; Thirion, B.; Grisel,
  O.; Blondel, M.; Prettenhofer, P.; Weiss, R.; Dubourg, V.; Vanderplas, J.;
  Passos, A.; Cournapeau, D.; Brucher, M.; Perrot, M.; and Duchesnay, E. 2011.
\newblock Scikit-learn: Machine Learning in {P}ython.
\newblock \emph{Journal of Machine Learning Research} 12: 2825--2830.

\bibitem[{Qiu, Li, and Zhao(2018)}]{Qiu}
Qiu, H.; Li, R.; and Zhao, J. 2018.
\newblock Dynamic Pricing in Shared Mobility on Demand Service.
\newblock \emph{arXiv: Optimization and Control} .

\bibitem[{Rebman(1974)}]{transportation-TU}
Rebman, K.~R. 1974.
\newblock Total unimodularity and the transportation problem: a generalization.
\newblock \emph{Linear Algebra and its Applications} 8(1): 11--24.
\newblock ISSN 0024-3795.
\newblock \doi{https://doi.org/10.1016/0024-3795(74)90003-2}.
\newblock
  \urlprefix\url{https://www.sciencedirect.com/science/article/pii/0024379574900032}.

\bibitem[{Riley, Legrain, and Van~Hentenryck(2019)}]{riley2019}
Riley, C.; Legrain, A.; and Van~Hentenryck, P. 2019.
\newblock Column Generation for Real-Time Ride-Sharing Operations.
\newblock In Rousseau, L.-M.; and Stergiou, K., eds., \emph{Integration of
  Constraint Programming, Artificial Intelligence, and Operations Research},
  472--487. Springer International Publishing.

\bibitem[{Riley, Van~Hentenryck, and Yuan(2020)}]{ijcai2020}
Riley, C.; Van~Hentenryck, P.; and Yuan, E. 2020.
\newblock Real-Time Dispatching of Large-Scale Ride-Sharing Systems:
  Integrating Optimization, Machine Learning, and Model Predictive Control.
\newblock In \emph{Proceedings of the Twenty-Ninth International Joint
  Conference on Artificial Intelligence, {IJCAI-20}}, 4417--4423. International
  Joint Conferences on Artificial Intelligence Organization.
\newblock \doi{10.24963/ijcai.2020/609}.
\newblock \urlprefix\url{https://doi.org/10.24963/ijcai.2020/609}.

\bibitem[{Verma et~al.(2017)Verma, Varakantham, Kraus, and Lau}]{Aug}
Verma, T.; Varakantham, P.; Kraus, S.; and Lau, H.~C. 2017.
\newblock Augmenting Decisions of Taxi Drivers through Reinforcement Learning
  for Improving Revenues.
\newblock In \emph{ICAPS}.

\bibitem[{Zhang, Rossi, and Pavone(2016)}]{Zhang}
Zhang, R.; Rossi, F.; and Pavone, M. 2016.
\newblock Model predictive control of autonomous mobility-on-demand systems.
\newblock In \emph{2016 IEEE International Conference on Robotics and
  Automation (ICRA)}, 1382--1389.
\newblock \doi{10.1109/ICRA.2016.7487272}.

\end{thebibliography}

\end{document}